%% file: acl/acl_latex.tex
\pdfoutput=1
\documentclass[11pt]{article}

\usepackage[]{acl}

\usepackage{times}
\usepackage{latexsym}
\usepackage{float} 
\usepackage{array} 
\usepackage{soul} 
\usepackage{xcolor} 
\usepackage{longtable}
\usepackage[T1]{fontenc}

\usepackage[utf8]{inputenc}

\usepackage{microtype}

\usepackage{inconsolata}

\input{math_commands.tex}

\usepackage{hyperref}
\usepackage{url}
\usepackage[linesnumbered,ruled,vlined]{algorithm2e}
\SetKw{KwInput}{Input}
\SetKw{KwInitialize}{Initialize}
\SetKwComment{Comment}{/* }{}
\SetKwInput{kwInit}{Init}
\usepackage{graphicx}
\usepackage{subcaption}
\usepackage{natbib}
\usepackage{doi}
\usepackage{amssymb}
\usepackage{multirow}
\usepackage{booktabs,adjustbox}
\usepackage{wrapfig,lipsum,booktabs}
\usepackage{enumitem}
\usepackage{xcolor}
\definecolor{lightgray}{gray}{0.3}

\title{Controllable Pareto Trade-off between Fairness and Accuracy}

\author{Yongkang Du$^{1}$, Jieyu Zhao$^1$, Yijun Yang$^{2}$, Tianyi Zhou$^3$\\
$^1$University of Southern California;$^2$University of Technology Sydney\\
$^3$University of Maryland, College Park\\}

\definecolor{mydarkred}{RGB}{139, 0, 0}

\begin{document}
{\makeatletter\acl@finalcopytrue
  \maketitle
}
\begin{abstract}
The fairness-accuracy trade-off is a key challenge in NLP tasks. Current work focuses on finding a single ``optimal'' solution to balance the two objectives, which is limited considering the diverse solutions on the Pareto front.
This work intends to provide controllable trade-offs according to the user's preference of the two objectives, which is defined as a reference vector. 
To achieve this goal, we apply multi-objective optimization (MOO), which can find solutions from various regions of the Pareto front. 
However, it is challenging to precisely control the trade-off due to the stochasticity of the training process and the high dimentional gradient vectors.
Thus, we propose \textbf{C}ontrollable \textbf{P}areto \textbf{T}rade-off (CPT) that can effectively train models to perform different trade-offs according to users' preferences.
CPT 1) stabilizes the fairness update with a moving average of stochastic gradients to determine the update direction, and 2) prunes the gradients by only keeping the gradients of the critical parameters. 
We evaluate CPT on hate speech detection and occupation classification tasks. Experiments show that CPT can achieve a higher-quality set of solutions on the Pareto front than the baseline methods. It also exhibits better controllability and can precisely follow the human-defined reference vectors.
\end{abstract}

\section{Introduction}

As language models (LMs) have shown human-level performance on various kinds of tasks, the fairness of LMs over different groups becomes a critical concern in practical applications. 
Unfairness in LMs can manifest in various ways and across different domains, e.g., LMs trained on biased text corpora can exhibit gender bias in text generation tasks~\citep{wan2023kelly,wambsganss2023unraveling}, encode societal stereotypes and prejudices present in the training data~\citep{charlesworth2022word,omrani2023social}.
Achieving fairness at the group level aim to emphasize that algorithmic decisions neither favor nor harm certain subgroups defined by the sensitive attribute, such as gender, race, religion, age, sexuality, nationality, and health conditions~\citep{chu2024fairness}. 

Current methods for group fairness can be divided into three categories ~\citep{hort2022bia,gallegos2023bias}.  
1) Pre-processing, which aims to balance the training data and prevent unfairness from affecting LMs~\citep{qian2022perturbation,garimella2022demographic}; 
2) In-processing, which optimizes the fairness loss function during the training process ~\citep{pmlr-v202-zhao23a}. The most intuitive strategy might be minimizing a linear combination of fairness and task loss~\citep{roy2022learning}. Another strategy is constrained optimization, which minimizes the task loss~\cite{cheng2022toward} under a fairness constraint; 
3) Post-processing, which aims at modifying the LMs's output to achieve group fairness~\citep{dhingra2023queer,pleiss2017fairness}. Although the above methods are developed to balance fairness and accuracy, it is still an open challenge for them to precisely control and customize the trade-off. 

Pre-processing and post-processing methods may roughly generate models with different preferences, but the effects are limited since they neglect the training process which is usually sensitive and complicated. 
In this paper, we focus on in-processing methods and intend to train a set of ``optimal'' models on the Pareto front, which is a set of equilibrium on which one cannot improve an objective without degrading another. 
While the Pareto front of fairness and accuracy can be highly complicated and contains rich solutions performing different trade-offs, simply combining objectives with different weights cannot guarantee visiting all of them and in the worst case, it may only end up with models optimized for a single objective (\S 4.7.4 of ~\citet{boyd_vandenberghe_2004}). 
Moreover, the conflicts between objectives or constraints can limit the exploration of diverse solutions on the Pareto front. 
As a result, models struggle to balance multiple objectives.

Multi-objective optimization (MOO) methods such as Multi-Gradient Descent Algorithm (MGDA) are able to converge to a Pareto equilibrium~\citep{desideri2012multiple} by finding a common optimization direction in each step on which all objectives are improving or at least staying the same. 
Moreover, given a pre-defined reference vector that indicates the preference for different objectives, MGDA has the potential to visit different regions of the Pareto front in the objective space. 
  
However, it is still challenging to directly apply MGDA to fairness-accuracy trade-off when training language models because:
1) MGDA relies on the full gradients of objectives to determine the common optimization direction, while stochastic gradient is more commonly used in training neural networks. Although stochasticity is important to generalization ability, it may lead to a drift of the fairness loss since samples in a mini-batch might not cover all subgroups. 
2) The inner products between gradients play an important role in determining the common optimization direction. However, when applied to train language models with millions of parameters, the curse of dimensionality might lead to less informative inner products reflecting the objective correlation. Moreover, many parameters can be pruned without affecting the model performance but they together may contaminate the inner product and thus are detrimental to the search for the common descent direction. 
3) It is challenging to control MGDA's optimization trajectory precisely following a pre-defined reference vector.

To overcome these challenges, we propose \textbf{C}ontrollable \textbf{P}areto Fairness-Accuracy \textbf{T}rade-off method (CPT). Our contribution can be summarized as follows: 
\begin{itemize}
    \item We utilize the moving average of stochastic gradients for each objective to approximate the full gradients used in MGDA for finding the common descent direction without missing subgroups. 
    \item We prune the gradient per objective and use a joint mask to reduce all gradients' dimensionality so MGDA can estimate a more precise common descent direction out of the pruned gradients.
    \item Our experiments on hate speech detection and occupation classification tasks show that CPT, compared to a rich class of baselines, can better follow the reference vectors and find diverse Pareto solutions with different trade-offs, resulting in a better hypervolume on the test set.
\end{itemize}

\section{Related Work}
\label{related_work}

\paragraph{Fairness-Aware Training}
Recently, more and more attention has been paid to fairness-aware training in natural language generation ~\citep{xu2022does, gupta2022mitigating}, natural language processing ~\citep{sheng2021societal}, and multi-task learning ~\citep{roy2022learning, oneto2019taking}. 
Commonly used methods can be divided into three categories. 
1) Regularization, which adds a term to the loss function to encourage the model's output independent with sensitive attributes. The regularization term mainly focuses on embedding~\citep{10.1609/aaai.v37i9.26279}, 
attention layers~\citep{attanasio-etal-2022-entropy}, and predicted token distribution~\citep{garg2019counterfactual}. 2) Constraint optimization, which sets an upper bound for unfairness that cannot be breached during training ~\citep{kim2018fairness, cheng2022toward, celis2019classification}. 3) Adversarial training, which simultaneously trains classification models and their discriminators (another classifier to predict sensitive attributes)~\citep{lahoti2020fairness,beutel2017data,han2022towards}. 

A significant fact for fairness-aware training is the trade-off between fairness and model performance. 
~\citet{huang2019stable} studies demographic parity and algorithmic stability from a theoretical perspective. 
~\citet{dutta2020there} investigates the essential trade-off between fairness and accuracy metrics. 
~\citet{liu2022accuracy} treats fairness as another objective, which is defined by the correlation between sensitive attributes and prediction results, and optimizes fairness and accuracy simultaneously. 
~\citet{tang2023theoretical} gives fine-grained categories to study properties of fairness-accuracy Pareto front. 
Here, we focus on the controllability of fairness-accuracy trade-off. 

\paragraph{Multi-Objective Optimization}
Multi-objective optimization (MOO) aims at optimizing more than one objective function. For a nontrivial MOO problem, no single solution exists that optimizes all objectives simultaneously. Instead, we strive to find a diverse set of solutions, each performing a different level of trade-off between these objectives. 
Existing methods for MOO can be divided into gradient-free methods and gradient-based methods. 

As for the gradient-free method, 
genetic algorithms~\citep{deb2002fast} and evolution strategies~\citep{coello2006evolutionary,deb2011multi} are commonly used for multi-objective optimization. 
However, it could be time-consuming to find the optimal solution especially when the searching space becomes large such as the parameters in the neural network. Also, the convergence cannot be guaranteed. 

As for the gradient-based method, the multi-gradient descent algorithm (MGDA) ~\citep{desideri2012multiple} applies gradient descent for solving MOO problems and is proved to converge to the Pareto stationary solution. 
However, MGDA suffers from expensive gradient computation and limited sparse solution issues. 
To address the first limitation, the Stochastic MultiSubgradient Descent Algorithm (SMSGDA) ~\citep{poirion2017descent} is purposed and applied in fairness-accuracy trade-off problem ~\citep{liu2022accuracy}. 
As for the second limitation, a branch of work utilizes the reference vector to guide the optimization and generate diverse solutions. 
Pareto multi-task learning (PMTL) ~\citep{lin2019pareto} divides the MOO problem into multiple subproblems according to the reference vectors and finds solutions in the regions that are close to reference vectors. 
Exact Pareto Optimal Search (EPO) ~\citep{mahapatra2020multi} is able to follow the reference vector more precisely by optimizing uniformity. 
WC-MGDA~\citep{momma2022multi} formulates the weighted Chebyshev problem to incorporate preference and derives MGDA-like components in the problem. 

However, limited work has been done to study gradient-based methods in the fairness-accuracy trade-off. 
In this paper, we present a novel method CPT, which applies MOO to achieve a controllable trade-off between fairness and accuracy. \looseness-1
\section{Method}
\label{method}
In this section, we define fairness-accuracy trade-off as a MOO problem in Section~\ref{sec: fairness-accuray trade-off}, introduce the key components of CPT from Section~\ref{sec:gradient moving average} to Section~\ref{sec: reference vector}, and give a detailed version of CPT in Section~\ref{sec: detailed-CPT}.
\subsection{Fairness-Accuracy Trade-off as MOO}
\label{sec: fairness-accuray trade-off}
MOO aims at optimizing multiple objectives simultaneously, which can be defined as below.
\begin{equation}
    \min _{{\bm{\theta}}} {\mathcal{L}}({\bm{\theta}}) \triangleq
    \left(\mathcal{L}_1({\bm{\theta}}), \mathcal{L}_2({\bm{\theta}}), \cdots, \mathcal{L}_m({\bm{\theta}})\right)^{\intercal}
\end{equation}
where $m$ is the number of objectives, ${\bm{\theta}}$ denotes the parameters to be optimized, $\mathcal{L}_i$ denotes the $i$-th objective.
Instead of finding one single solution in general single-objective optimization, we strive to achieve Pareto stationarity in MOO.

\paragraph{Definition 1 \quad Pareto stationarity for MOO}
A solution ${\bm{\theta}}^{*}$ is Pareto stationary if there exist $\boldsymbol{\alpha} \in \mathbb{R}^m$ such that 
\text{$\sum_{i=1}^{m}\alpha_{i}=1$},
\text{$\alpha_{i}\geq 0$}, and 
\text{$\sum_{i=1}^{m}\alpha_{i}\nabla_{{\bm{\theta}}}\mathcal{L}_{i}({\bm{\theta}}^{*})=\boldsymbol{0}$}, 
which implies that we cannot find a common updating direction for improving all objectives. \looseness-1

The Pareto front \text{\small$\mathcal{P}$} represents a set of Pareto stationary solutions, in which each solution achieves a certain trade-off between the objectives. 
\text{\small$\mathcal{P}$} forms a boundary in the objective space, and any point inside this boundary represents a suboptimal solution because it can be improved in at least one objective without degrading others. 

\paragraph{Definition 2\quad Fairness-Accuracy Trade-off}

Given a dataset \text{$D$} with \text{$n$} samples, consisting of input features $X$, labels $Y$ (\text{$c$} number of classes), sensitive attributes \text{$A$} (such as the demographic group information), and a classifier $f$ parameterized by \text{${\bm{\theta}}$}, we utilize CrossEntropy for classification loss \text{$\mathcal{L}_{acc}$}, which is defined by Eq.~\ref{eql:accloss}.
\begin{equation}
    \mathcal{L}_{acc} = -\frac{1}{n}\sum_{i=1}^{n}\sum_{j=1}^{c}{y_{ij}\log(f(x_{ij}))}
\label{eql:accloss}
\end{equation}
where $y_{ij}$ and $f(x_{ij})$ indicate the label and the prediction of $j$-th class of $i$-th sample respectively.  
We utilize DiffEodd~\citep{chuang2021fair} for fairness loss \text{$\mathcal{L}_{fair}$} (defined by Eq.~\ref{eql:fairloss}), which is the gap regularization method for equalized odd (EODD)~\citep{hardt2016equality}. 
For each \text{$y \in Y$}, DiffEodd intends to minimize the gap between the conditional prediction probability given a certain attribute \text{$a$} and the overall prediction probability (see Eq.~\ref{eql:prob}).
\begin{equation}
    \begin{gathered}
        P_{all}^y = P(f(X)|Y=y),\\P_a^y = P(f(X)|A=a, Y=y)
    \end{gathered}
    \label{eql:prob}
\end{equation}
\begin{equation}
  \mathcal{L}_{fair}=  \sum_{a \in A}\sum_{y \in Y}\left|\mathbb{E}_{x \sim P_a^y} f(x)-\mathbb{E}_{x \sim P_{all}^y} f(x)\right|
    \label{eql:fairloss}
\end{equation}
Our goal is to train $f$, so that it can perform well on classification tasks and make fair predictions for each subgroup. 
Then, the fairness-accuracy trade-off problem could be defined as:
\begin{equation}
    \min _{\bm{\theta}} \mathcal{L}({\bm{\theta}}) 
\triangleq \left(\mathcal{L}_{fair}({\bm{\theta}}), \mathcal{L}_{acc}({\bm{\theta}})\right)^{\intercal}
    \label{objective function}
\end{equation}
\paragraph{Definition 3 \quad Common Descent Vector}

When using the gradient-based optimization algorithm to solve the MOO problem, the common decent vector ${g}$ provides the direction for optimization and the distance to update along the direction.
MGDA defines the common descent vector as the vector with minimum L2 norm in the convex hull of the gradient of each objective (see Eq.~\ref{eql:minnorm}) so that the objectives will not conflict with each other. 
\begin{equation}
    \min _{\boldsymbol{\alpha}}\left\|\boldsymbol{\alpha} \nabla_{\bm{\theta}} \mathcal{L}\left({\bm{\theta}}\right)\right\|_2, \text { s.t. }\|\boldsymbol{\alpha}\|_1=1, \boldsymbol{\alpha} \geq \mathbf{0}
    \label{eql:minnorm}
\end{equation}
where $\nabla_{\bm{\theta}} \mathcal{L}({\bm{\theta}})$ indicates the gradient of the objective function and $\boldsymbol{\alpha}$ is the combination weights. 
Eq.~\ref{eql:minnorm} can be solved with the Frank-Wolfe algorithm~\citep{DBLP:conf/icml/Jaggi13}. Then the common descent vector in multi-objective optimization can be defined as:
\begin{equation}
    {g} \triangleq\sum_{i=1}^{m}\alpha_{i}\nabla_{{\bm{\theta}}}\mathcal{L}_{i}({\bm{\theta}})
    \label{eql:common_descent_vector}
\end{equation}

In this paper, we extend MGDA to fairness-accuracy trade-off and propose a novel method named CPT to generate controllable Pareto stationary solutions.

\subsection{Moving Average of Stochastic Gradient to Address Fairness Loss Drift}
\label{sec:gradient moving average}
When optimizing one single objective, we usually employ a stochastic approach, where a subset of data is used to compute the mini-batch stochastic gradient. 
However, directly using stochastic gradients for MOO may not be a wise choice. 
First, the stochastic nature of the optimization process introduces noise into the gradient, which could be misleading for calculating the common descent vector. 
Second, as one single mini-batch may not cover all the subgroups, the mini-batch fairness loss as well as its gradient could be inaccurate. 

Inspired by SGD with momentum, which intends to stabilize the gradient during optimization, CPT keeps moving average gradients (see Eq.~\ref{eql:gradavg}) to approximate the whole gradients of objective functions. 
This method smooths the gradient of each objective before calculating the common descent vector, which leads to a more precise weight for each objective. 
Also, by accumulating the previous fairness gradients, CPT takes into account those subgroups that might be missing in the current mini-batch, which leads to a better fairness goal. 

The moving average gradient of step $k$ is calculated with:
\begin{equation}
    \bar{G}^{k} = \beta*\bar{G}^{k-1}+
    (1-\beta)*\nabla_{{\bm{\theta}}}\mathcal{L}({\bm{\theta}})
    \label{eql:gradavg}
\end{equation}
where $\Bar{G}$ and $\beta$ are the moving average gradient and the moving average weight. 

\subsection{Gradient Pruning in MGDA}

In addition to refining MGDA with moving average gradient in Section~\ref{sec:gradient moving average}, we also intend to get a better common descent vector by denoising the gradient vector and lowering its dimension. 

When searching the direction for the common descent vector, MGDA uses inner products of gradient vectors (more details in Appendix~\ref{sec: common descent vector}). 
However, high-dimension gradients could be dominated by noise, making the common descent vector calculated by MGDA imprecise. 
Since the parameters with higher values are more influential for the optimization process, we generate a mask based on the parameters’ magnitude and filter out the gradients of parameters with low magnitude. 
The pruning mask $\mathbf{M}$ is initialized as a matrix of ones who has the same shape as ${\bm{\theta}}$.
Then we apply Alg.~\ref{alg:magnitude_pruning} to generate the pruning mask $\mathbf{M}$. Given the parameter ${\bm{\theta}}$ of the neural network and the pruning ratio $\gamma$, we first compute the average magnitude of ${\bm{\theta}}$ and get the pruning threshold $\gamma \| {\bm{\theta}} \|_1$. Then we iterate the parameter and generate the corresponding pruning mask.
\begin{algorithm}[t]
    \caption{Generation of Pruning Mask}
    \label{alg:magnitude_pruning}
    \KwInput parameter ${{\bm{\theta}}}$, pruning mask $\mathbf{M}$, pruning ratio $\gamma$\\
    \For{${\theta} \in {{\bm{\theta}}}, M \in \mathbf{M}$}
    {
        \If{$|{\theta}| \leq \gamma \|{{\bm{\theta}}} \|_1$}{
            $M\leftarrow 0$
        }
    }
    \textbf{Output} Pruning mask $\mathbf{M}$
\end{algorithm}
The pruned gradient is calculated by:
\begin{equation}
    \widetilde{G}_{i} = \mathbf{M} \odot \bar{G}_{i}
    \label{eql:gradprun}
\end{equation}
where $\Bar{G}_{i}$ is the moving average gradient of objective $i$. 
With gradient pruning, we are able to accelerate the computation as well as get a better common descent vector. 
In order to keep the theoretical guarantee of MGDA (the objectives will not confilict with each other), we apply the pruned gradient to calculate the combination weights $\boldsymbol{\alpha}$ and the common descent vector $g$ (Eq.~\ref{eql:common_descent_vector} is updated by Eq.~\ref{eql:pruned_common_descent_vector}) and only update the parameters that have non-zero gradients.  
\begin{equation}
    {g} \triangleq
    \sum_{i=1}^m \alpha_i \widetilde{G}_{i}
    \label{eql:pruned_common_descent_vector}
\end{equation}
\subsection{Reference Vector Following}
\label{sec: reference vector}
To better control the optimization, CPT utilizes reference vector \text{$\vec{v}=\left(v_{fair},v_{acc}\right)$}, where \text{$v_{fair}, v_{acc} \in \mathbb{R}$}, to guide the optimization process.
The reference vector indicates the expected ratio of two loss values. When \text{\small$\frac{v_{fair}}{v_{acc}} > 1$}, we except $\mathcal{L}_{acc}$ to be lower than $\mathcal{L}_{fair}$, which means a preference for accuracy.
We define the constraint loss by the Kullback–Leibler (KL) divergence between the loss value vector $\vec{l}$ and the reference vector $\vec{v}$ (see Eq.~\ref{eql: constraint bi-objective optimization}).
Different reference vectors set different constraints for the optimization process and lead to diverse trade-offs on the Pareto set. 
\begin{equation}
    \Psi(\vec{l},\vec{v}) \triangleq D_{KL}\left(\frac{\vec{l}}{\lVert \vec{l} \rVert_{1}} || \frac{\vec{v}}{\lVert \vec{v} \rVert_{1}}\right),
    \label{eql: constraint bi-objective optimization}
\end{equation}
where \text{\small$\vec{v}=\left(v_{fair},v_{acc}\right)$} is the reference vector and \text{\small$\vec{l}=\left(\mathcal{L}_{fair}, \mathcal{L}_{acc}\right)$} is the vector for two objective loss values. 

CPT applies two stages for the optimization: correction stage and MOO stage.
In the correction stage, CPT applies single objective optimization to satisfy the constraint: \text{\small$\Psi(\vec{l},\vec{v}) < \psi$},
where $\psi$ is a predefined threshold. 
The correction stage provides a suitable starting point for the MOO stage that follows the reference vector $\vec{v}$. 
In the MOO stage, CPT simultaneously optimizing three objectives including fairness loss $\mathcal{L}_{fair}$, classification loss $\mathcal{L}_{acc}$, and the constraint loss $\Psi(\vec{l},\vec{v})$. 
Thus, the objective function for MOO stage can be written as:
\begin{equation}
    \min _{\bm{\theta}} \mathcal{L}({\bm{\theta}}) \triangleq 
    \min _{\bm{\theta}} \left(\mathcal{L}_{fair}({\bm{\theta}}), \mathcal{L}_{acc}({\bm{\theta}}),
    \Psi(\vec{l},\vec{v})\right)^{\intercal}
    \label{eql:tri_objective}
\end{equation}

\subsection{Controllable Pareto Fairness-Accuracy Trade-off}
\label{sec: detailed-CPT}
In this section, we provide a detailed version of CPT in Alg.~\ref{alg:cpt}, which generates a controllable Pareto fairness-accuracy trade-off. 
CPT first finds a starting point for multi-objective optimization that satisfies the constraint set by the reference vector $\vec{v}$ in the correction stage. 
Then it jointly optimizes fairness loss $\mathcal{L}_{fair}$, classification loss $\mathcal{L}_{acc}$, and the constraint loss $\Psi(\vec{l},\vec{v})$ to find the Pareto stationary solution in a certain region in the MOO stage. 

The moving average gradients for accuracy $\Bar{G}_{acc}$ and fairness $\Bar{G}_{fair}$ are updated through the whole optimization process, while $\Bar{G}_{kl}$ is updated only in the MOO stage.  
Meanwhile, CPT prunes the moving average gradient of each objective with the mask $\mathbf{M}$. Finally, CPT computes the common descent vector $g$ and updates the parameters. 

\looseness-1

\begin{algorithm}[t]
\caption{Training Procedure of CPT}
\label{alg:cpt}
\KwInput dataset $D=\{(X,Y,A)\}$, reference vector $\vec{v}=(v_{{\text{fair}}},v_{{\text{acc}}})$, threshold $\psi$, moving average weight $\beta$, pruning ratio $\gamma$, learning rate $\eta$\\
\KwInitialize model $f({\bm{\theta}})$, FrankWolfeSolver $\mathtt{F}$~\citep{DBLP:conf/icml/Jaggi13}, $\bar{G}_{\text{fair}},\bar{G}_{\text{acc}},\bar{G}_{\text{KL}}\leftarrow\mathbf{0},\mathbf{0},\mathbf{0}$\\

\For{$k=0,\dots,K$}
{
Get pruning mask $\mathbf{M}_{k}$ by Alg.~\ref{alg:magnitude_pruning} with $\gamma$\\
Get $\mathcal{L}^{k}_{\text{acc}}$ and $\mathcal{L}^{k}_{\text{fair}}$ by Eq.~(\ref{eql:accloss}) and Eq.~(\ref{eql:fairloss})\\
\Comment{\textcolor{mydarkred}{Gradient Moving Average}}
Update $\bar{G}_{\text{fair}}$ and $\bar{G}_{\text{acc}}$ by Eq.~(\ref{eql:gradavg}) with $\beta$\\
\If{$\Psi(\bm{\mathcal{L}},\vec{v})>\psi$}
{ 
    \Comment{\textcolor{mydarkred}{Correction Stage}}
    \If{$\mathcal{L}_{\text{\rm fair}}/\mathcal{L}_{\text{\rm acc}} > v_{\text{\rm fair}}/v_{\text{\rm acc}}$}{
        Get descent direction $g = \bar{G}_{\text{fair}}$
    }
    \Else{
        Get descent direction $g = \bar{G}_{\text{acc}}$
    }
}
\Else{
    \Comment{\textcolor{mydarkred}{MOO Stage}}
    Update $\bar{G}_{\text{KL}}$ by Eq.~(\ref{eql:gradavg}) with $\beta$\\

    \Comment{\textcolor{mydarkred}{Gradient Pruning}}
    Get $\tilde{G}_{\text{fair}}, \tilde{G}_{\text{acc}}, \tilde{G}_{\text{KL}}$ by Eq.~(\ref{eql:gradprun}) with $\mathbf{M}_{k}$\\
    \Comment{\textcolor{mydarkred}{Compute Combination Weights}}
    $\boldsymbol{\alpha}_{k}=\mathtt{F}(\tilde{G}_{\text{fair}}, \tilde{G}_{\text{acc}}, \tilde{G}_{\text{KL}})$\\
    Get descent vector $g$ by Eq.~(\ref{eql:pruned_common_descent_vector}) with $\boldsymbol{\alpha}_{k}$\\
}
\Comment{\textcolor{mydarkred}{Parameter Update}}
${\bm{\theta}}_{t+1}={\bm{\theta}}_{t}-\eta {g}$
}
\textbf{Output} Pareto-optimal solution ${\bm{\theta}}^{*}$ following $\vec{v}$
\end{algorithm}

\section{Experiments}
\label{experiment}
In this section, we evaluate CPT from the following aspects. 1) Can CPT control the fairness-accuracy trade-off by precise reference vector following? 2) Can CPT generate more diverse trade-off solutions between the two objectives? 3) Can the trade-off solution obtained by CPT generalize to unseen data? Specifically, Section~\ref{sec: expset} describes the experimental setting. Section~\ref{sec:result} shows the superiority of CPT by comparing it with several state-of-the-art (SoTA) MOO methods. Section~\ref{sec:ablation} presents a thorough ablation study to demonstrate the effectiveness of gradient moving average and gradient pruning. We show the result of the case study in Apendix~\ref{sec: casestudy}.

\subsection{Experimental Setting}
\label{sec: expset}
\paragraph{Benchmarks }
 
We use Jigsaw dataset \footnote{https://www.kaggle.com/competitions/jigsaw-unintended-bias-in-toxicity-classification/data} to evaluate CPT on the toxicity classification task and focus on race bias as it has been proved to show the most significant bias over other attributes~\citep{cheng2022toward}. 
In addition, we use BiasBios dataset~\citep{de2019bias} to evaluate CPT on the occupation classification task and focus on gender bias. Following~\citet{brandl2023interplay}, we use a subset of the original dataset which contains five medical occupations with clear gender imbalance. The statistics of the datasets are shown in Appendix~\ref{sec:data-stat}.

We utilize accuracy as the classification metric and EODD as  the fairness metric to evaluate CPT. 
The fairness metric is the difference between true positive rate (TPR) and false positive rate (FPR) under different sensitive attributes and the overall TPR and FPR (see Eq.~\ref{eql:fairmetric}).
\begin{equation}
    \sum_{a \in \mathbf{A}}\left(
    \left|\mathrm{TPR}_{a}-\mathrm{TPR}_{\text {overall }}\right|+
    \left|\mathrm{FPR}_{a}-\mathrm{FPR}_{\text {overall }}\right|
    \right)
    \label{eql:fairmetric}
\end{equation}
A higher accuracy indicates better classification performance and a lower EODD value indicates there is less bias among predictions of different subgroups. 

\paragraph{Baselines} We compare CPT with several baselines and SoTA MOO methods below: \\
(1) {\bf Scalarization} that directly optimizes a weighted sum of multiple objectives. \\
(2) {\bf MGDA~\citep{sener2018multi} with diverse initialization}: We first provide MGDA with diverse initial solutions and then apply MGDA to solve the multi-objective optimization problem with respect to each of them. \\
(3) {\bf Pareto Multi-Task Learning (PMTL)}~\citep{lin2019pareto} generates solutions falling to different regions of the Pareto front by decomposing a multi-objective optimization problem into multiple sub-problems, each characterized by a distinct preference among those objectives. \\
(4) {\bf Exact Pareto Optimization (EPO)}~\citep{mahapatra2020multi} combines multiple gradient descent with an elaborate projection operator to achieve convergence to the required Pareto solution.\\
(5) {\bf CPT(w/o Prune)}: CPT without gradient pruning.\\
(6) {\bf CPT(w/o GA)}: CPT without gradient moving average.

\paragraph{Training details} We apply sentence transformer~\citep{reimers2019sentence} as the encoder and stack two fully connected layers as classification heads. 
We use an SGD optimizer with an initial learning rate of 0.01, which is decayed by a small constant factor of 0.8 until the number of epochs reaches a pre-defined value. All of the experiments are conducted on a 4090Ti GPU with four random seeds for fair comparison. More details on the hyperparameters used in training can be found in the Appendix~\ref{sec:train-details}. In order to represent different trade-offs between fairness and accuracy, we set a diverse set of reference vectors: $\sV = \{(2,1), (3,2), (1,1), (2,3), (1,2), (1,3)\}$.
By optimizing the loss function with the chosen reference vector (see Eq.~\ref{eql: constraint bi-objective optimization}), CPT can precisely control the trade-off between fairness and accuracy.

\begin{figure}[t]
\centering
  \begin{subfigure}{0.49\linewidth} 
    \centering
    \includegraphics[width=\linewidth]{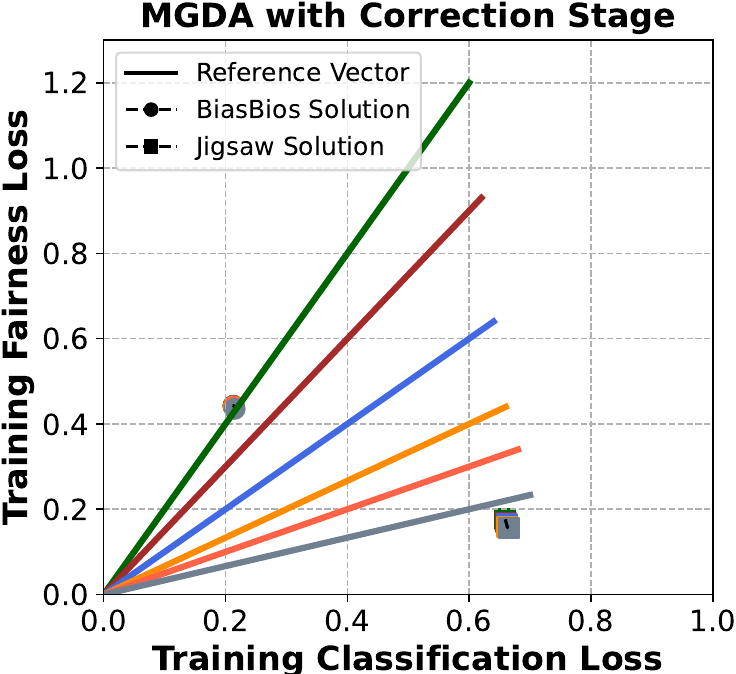}
    \caption{}
    \label{fig:refmgda}
  \end{subfigure}
  \begin{subfigure}{0.49\linewidth} 
    \centering
    \includegraphics[width=\linewidth]{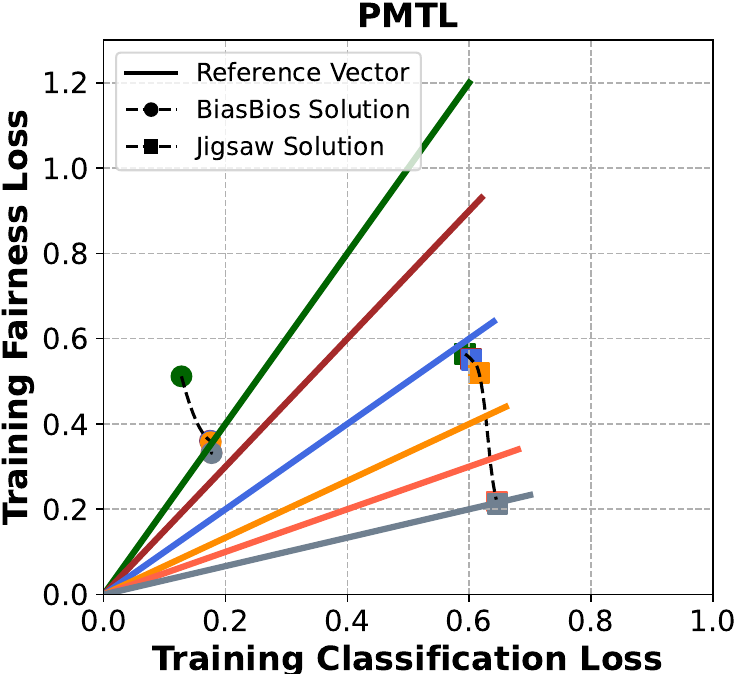}
    \caption{}
    \label{fig:refpmtl}
  \end{subfigure}
  
  \begin{subfigure}{0.49\linewidth} 
    \centering
    \includegraphics[width=\linewidth]{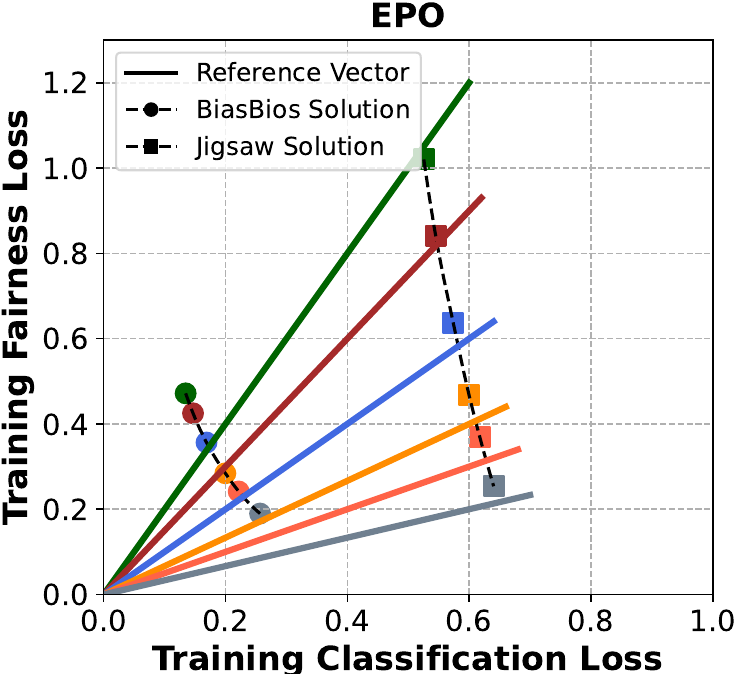}
    \caption{}
    \label{fig:refepo}
  \end{subfigure}
  \begin{subfigure}{0.49\linewidth} 
    \centering
    \includegraphics[width=\linewidth]{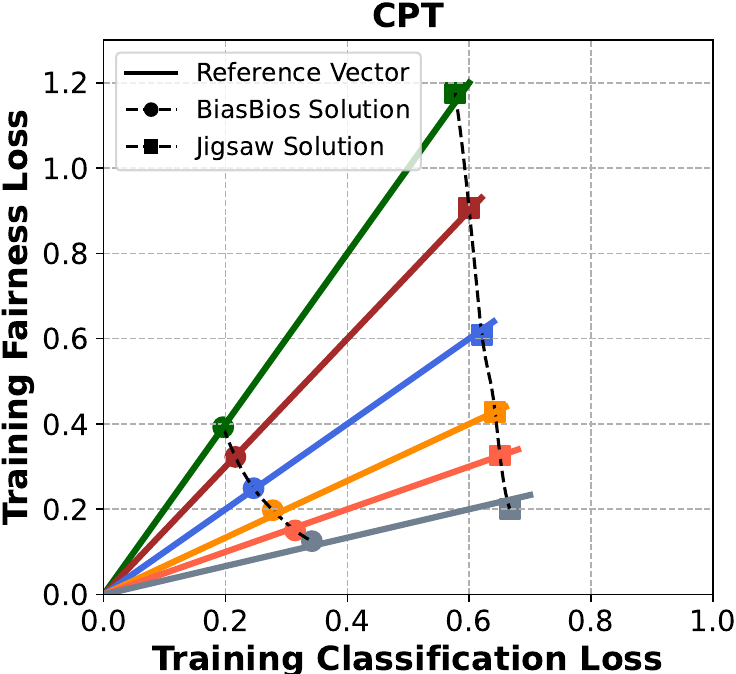}
    \caption{}
    \label{fig:refcpt}
  \end{subfigure}
    
  \begin{subfigure}{0.49\linewidth} 
    \centering
    \includegraphics[width=\linewidth]{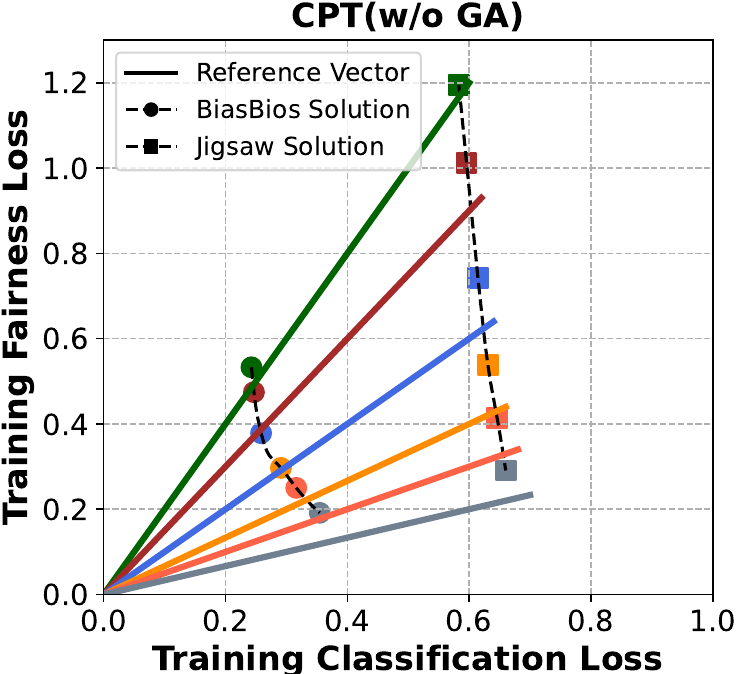}
    \caption{}
  \end{subfigure}
  \begin{subfigure}{0.49\linewidth} 
    \centering
    \includegraphics[width=\linewidth]{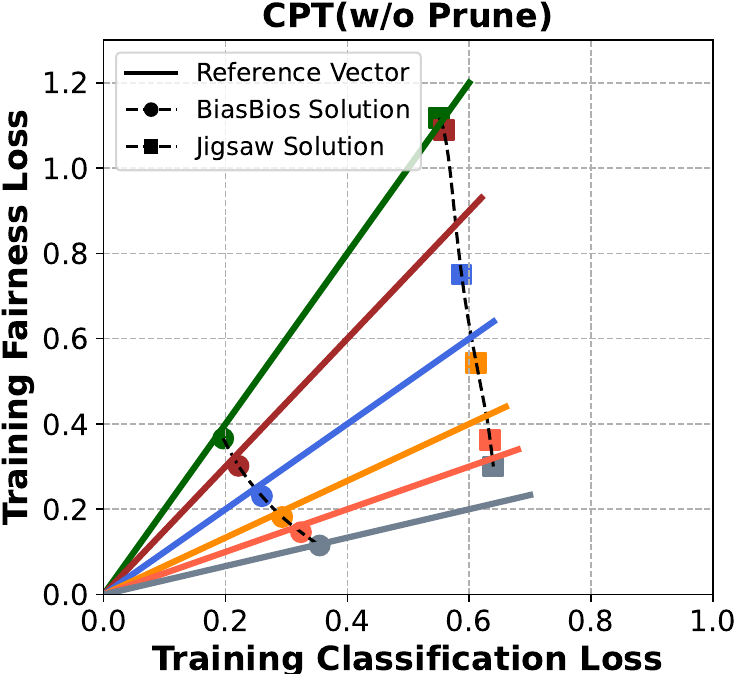}
    \caption{}
  \end{subfigure}
  \caption{Fairness-accuracy trade-off solutions generated by different methods using six reference vectors.  
  {\bf Among all methods, CPT (Figure~\ref{fig:refcpt}) is the best one whose solutions precisely follow the reference vectors}. Reference vectors from top to bottom are $(2,1), (3,2), (1,1), (2,3), (1,2), (1,3)$. The x-axis denotes the classification loss while the y-axis denotes the fairness loss on the training set.}
  \label{fig:minref}
\end{figure}

\subsection{Main Results}
\label{sec:result}
\paragraph{Controllable Pareto trade-off by following reference vector.} 
In order to demonstrate the advantage of CPT, we compare it with MGDA with diverse initialization (Figure~\ref{fig:refmgda}) and two SoTA reference vector-based MOO methods: PMTL (Figure~\ref{fig:refpmtl}) and EPO (Figure~\ref{fig:refepo}).
As shown in Figure~\ref{fig:minref}, MGDA can only generate one single solution even with different initialization. PMTL fails to generate diverse solutions with given reference vectors, and the solutions are mainly located in two regions. One possible explanation is that PMTL only uses reference vectors to determine initial solutions but lacks a principled method to follow them during the rest of the optimization process.

While EPO achieves lower accuracy and fairness loss values than CPT for vectors with a preference for the accuracy objective (see $\vec{v}=(2,1)$ and $\vec{v}=(3,2)$), this advantage disappears on unseen data. The results in Figure~\ref{fig:epo-cpt-j} and Figure~\ref{fig:hp(epo-cpt-bb)} (in Appendix) indicate that EPO achieves worse fairness performance on the testing set for reference vectors $\vec{v}=(2,1)$ and $\vec{v}=(3,2)$, reflecting that EPO suffers from overfitting to training data. 
Furthermore, EPO fails to follow the reference vectors with higher fairness preference. 
This is because EPO uses a noisy stochastic gradient to determine the update direction for each step, which could be inaccurate as we discussed in Section~\ref{sec:gradient moving average}, and thus the fairness performance is harmed. 
This challenge is successfully solved by CPT. 
Benefits from the pruning and moving average of gradients, CPT is able to precisely follow each reference vector. We show that the KL divergence between the reference vector and the loss value vector first decreases and then stays stable for the rest of the training process (see Figure~\ref{fig:training-kl-loss} in Appendix), which indicates that the training process is well-guided by the reference vector.

\begin{table*}
  \centering
  \small
  \renewcommand\arraystretch{1.1}
  \resizebox{\textwidth}{!}
  {%
  \begin{tabular}{cl cccccc c cccccc}
    \toprule
    \multirow{2}{*}{\rotatebox[origin=c]{90}{\textbf{\shortstack{Data\\-set}}}} & \multirow{2}{*}{\textbf{Method}}
     & \multicolumn{6}{c}{\textbf{Accuracy ($\uparrow$})} & &\multicolumn{6}{c}{\textbf{EODD ($\downarrow$})}\\
    \cmidrule{3-8}
    \cmidrule{10-15}
    & & (2,1) & (3,2) &(1,1)&(2,3)&(1,2)&(1,3) & 
    & (2,1) & (3,2) &(1,1)&(2,3)&(1,2)&(1,3)\\
    \midrule
    \multirow{7}{*}{\rotatebox[origin=c]{90}{\shortstack{Jigsaw}}} 
    & Scalarization
    &\textbf{{+0.48}}
    &{+0.07}
    &74.35
    &{-0.15}
    &{-0.30}
    &{-23.55}& 
    
    &{+1.54}
    &{+0.84}
    &5.86
    &{-0.73}
    &{-0.72}
    &\textbf{{-5.75}} \\
    &MGDA
    &{-1.13}
    &{-1.13}
    &\textbf{71.96}
    &{-0.26}
    &{-0.33}
    &{-0.51}& 
    
    &{+1.28}
    &{+1.28}
    &\textbf{8.20}
    &{+2.68}
    &{+4.30}
    &{+1.78}\\
    &PMTL
    &\textbf{{+0.67}}
    &{-4.09}
    &72.71
    &{-0.93}
    &{-0.54}
    &{-0.21}& 
    
    &{+0.37}
    &{-0.34}
    &4.90
    &{+4.37}
    &\textbf{{-1.02}}
    &{-0.32}\\
    &EPO
    &\textbf{{+0.31}}
    &{+0.80}
    &73.63
    &{+0.03}
    &{-0.25}
    &{-0.54}& 
    
    &{-1.77}
    &{-1.39}
    &6.69
    &{-1.61}
    &\textbf{{-1.85}}
    &{-1.15}\\
    &CPT(w/o Prune)
    &{+0.33}
    &\textbf{{+0.46}}
    &73.48
    &{-0.52}
    &{-1.52}
    &{-1.26}& 
    
    &\textbf{{-1.44}}
    &{+0.16}
    &5.64
    &{+0.12}
    &{+1.81}
    &{+1.24}\\
    &CPT(w/o GA)
    &\textbf{{+1.41}}
    &{+1.26}
    &71.55
    &{-0.80}
    &{-1.11}
    &{-2.01}& 
    
    &{+6.87}
    &{+5.26}
    &\textbf{1.74}
    &{+0.38}
    &{+2.65}
    &{+1.83}\\
    &CPT
    &\textbf{{+1.11}}
    &{+0.44}
    &72.09
    &{-0.70}
    &{-1.31}
    &{-2.29}& 
    
    &{+4.39}
    &{+1.81}
    &3.47
    &{-0.66}
    &{-0.89}
    &\textbf{{-0.92}}\\
    \midrule
    \multirow{7}{*}{\rotatebox[origin=c]{90}{\shortstack{BiasBios}}}
    & Scalarization
    &\textbf{+0.14} & +0.11 & 91.09 & -0.15 & -0.27 & -0.46 & 
    
    &+0.64 & +0.22 & 8.68 & -0.79 & -1.11 & \textbf{-1.30} \\
    &MGDA
    &\textbf{+0.02} & -0.03 & 90.49 & +0.01 & -0.06 & -0.04 &
    
    &+0.00 & +0.05 & \textbf{7.19} & +0.08 & +0.11 & +0.00\\
    &PMTL
    &\textbf{+0.88} & -0.2 & 90.24 & +0.06 & +0.23 & +0.26 &
    &+0.61 & +0.03 & +6.94 & \textbf{-0.079} & +0.00 & +0.01 \\
    &EPO
    &\textbf{+0.49} & +0.24 & 90.28 & -1.00 & -2.36 & -6.89 &
    &+0.94 & +0.5 & \textbf{7.01} & +0.35 & +0.19 & +1.17 \\
    &CPT(w/o Prune)
    &\textbf{+4.04}
    &{+2.47}
    &84.36
    &{-1.85}
    &{-3.38}
    &{-6.45}& 
    
    &{-0.01}
    &{-0.02}
    &7.31
    &{-0.23}
    &{-0.78}
    &\textbf{-2.34}\\
    &CPT(w/o GA)
    &\textbf{{+3.4}}
    &{+2.16}
    &84.54
    &{-3.19}
    &{-6.94}
    &{-12.01}& 
    
    &{-0.60}
    &{-0.48}
    &{7.50}
    &{-0.10}
    &{-2.56}
    &\textbf{-3.60}\\
    &CPT
    &\textbf{{+2.71}}
    &{+1.66}
    &85.31
    &{-1.87}
    &{-4.27}
    &{-9.53}& 
    
    &{+0.01}
    &{-0.09}
    &7.84
    &{-0.37}
    &{-0.89}
    &\textbf{{-3.59}}\\
    \bottomrule 
  \end{tabular}
  }
  \caption{\looseness=-1 Accuracy and EODD (fairness) trade-off on the test set. The results for reference vector $v=(1,1)$ are reported in their original values, while the results for the other five reference vectors are differences from metrics achieved at $v=(1,1)$. 
  For each method, the best accuracy and fairness among the six reference vectors are highlighted by \textbf{bold}.  
  {\bf CPT's fairness and accuracy on the test set better match the reference vectors} than other methods.} 
  \label{tab:testperf}
\end{table*}

\paragraph{Evaluate solutions' quality with fairness weighted hypervolume.}
We evaluate CPT on the testing set and show the result in Table~\ref{tab:hp} and Figure~\ref{fig:hypervolume} in Appendix. For a fair comparison, we apply the same reference point $(2, 1)$ for all methods.
Hypervolume~\citep{zitzler1999multiobjective} is a widely used metric in MOO. It calculates the area/volume of the resulting set of nondominated solutions with respect to a reference point to measure the diversity of these solutions (more details can be found in Appendix~\ref{sec: hypervolume}). 
In the experiment, the reference point is the worst-case result for each objective, i.e., the largest classification and fairness losses (the yellow point on the top right corner in Figure~\ref{fig:hypervolume}). 

However, the original hypervolume metric neglects the difficulty of optimization for different objectives and treats them equally. For example, in our case, the fairness loss is harder optimize than the classification loss. In order to address this issue, we utilize a reference point that is more favorable to fairness. As shown in Table~\ref{tab:hp}, CPT and CPT(w/o Prune) achieves the best performance compared with other methods.

\begin{table}[]
    \small
    \centering
    \begin{tabular}{lrr}
        \toprule
        \multirow{2}{*}{\textbf{Method}}
         & \multicolumn{2}{c}{\textbf{Hypervolume}} \\
        \cmidrule{2-3}
        & Jigsaw & BiasBios \\
        \midrule
        Scalarization & 0.53 & 0.29 \\
        MGDA & 0.63 & 0.33 \\
        PMTL & 0.69 & 0.45\\
        EPO & 0.72 & 0.53 \\
        CPT(w/o GA) & 0.70 & 0.51 \\
        CPT(w/o Prune) & 0.71 & \textbf{0.55} \\
        CPT & \textbf{0.73} & 0.54 \\
        \bottomrule
    \end{tabular}
    \caption{Hypervolume (test set) of the solutions achieved by different methods in the fairness-accuracy space. CPT and CPT(w/o Prune) achieve the best hypervolume on the test set.}
    \label{tab:hp}
\end{table}

\paragraph{Generalizable Pareto trade-off to unseen data.} 
When addressing the fairness-accuracy trade-off in real-world prediction problems, the resulting models are expected to work on training data meanwhile generalizing to unseen data. 
Hence, a reliable method should achieve a consistent fairness-accuracy trade-off on training and testing sets under the same reference vector. An ideal result in our experiment should satisfy: 1) Achieve the highest accuracy when $v=(2,1)$ and the lowest EODD when $v=(1,3)$; 2) For reference vectors with preference on accuracy (fairness), the results are expected to show higher (lower) accuracy and higher (lower) EODD than $v=(1,1)$. As shown in Table~\ref{tab:testperf}, only scalarization and CPT exhibit these characteristics. However, when $v=(1,3)$, the model trained with scalarization performs like a random model on the Jigsaw dataset.

\subsection{Ablation study}
\label{sec:ablation}
Here we study how the moving average and pruning of the objectives' gradients affect the performance. 
Comparing CPT(w/o GA) with CPT in Figure~\ref{fig:minref}, we find that there is a consistent increase of fairness loss for nearly all solutions, demonstrating that the gradient moving average technique can lead to a better fairness performance. 
On the other hand, when CPT removes the gradient pruning, the optimization process becomes more unstable, highlighting the importance of gradient pruning in stabilizing the optimization and determining a more accurate descent direction. 

\begin{figure}
\centering
  \begin{subfigure}{0.49\linewidth} 
    \centering
    \includegraphics[width=\linewidth]{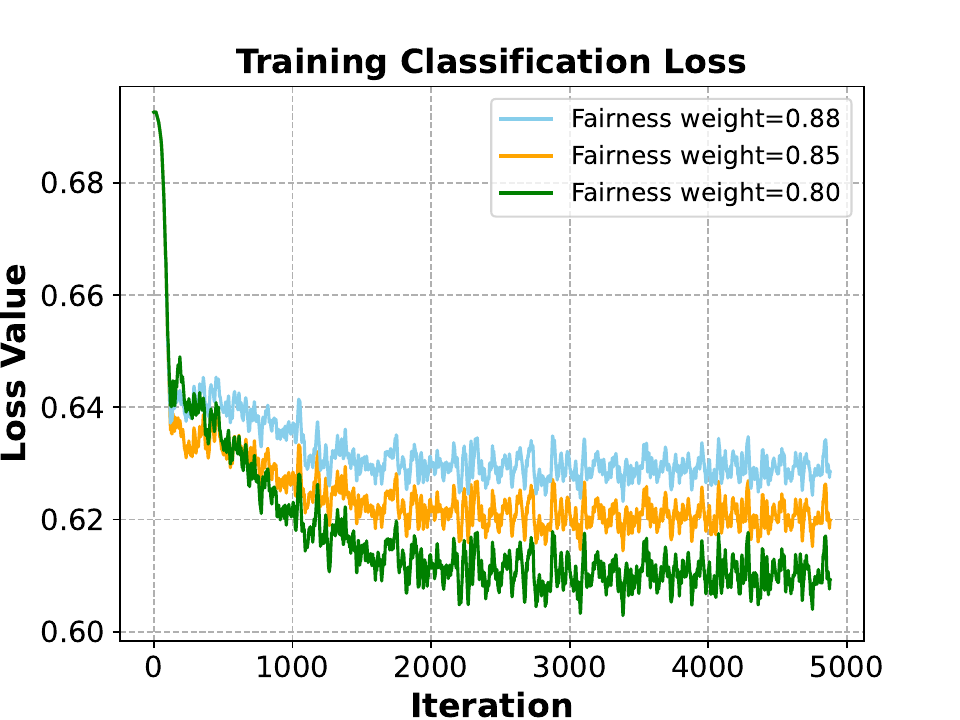}
    \caption{Classification Loss}
  \end{subfigure}
  \begin{subfigure}{0.49\linewidth} 
    \centering
    \includegraphics[width=\linewidth]{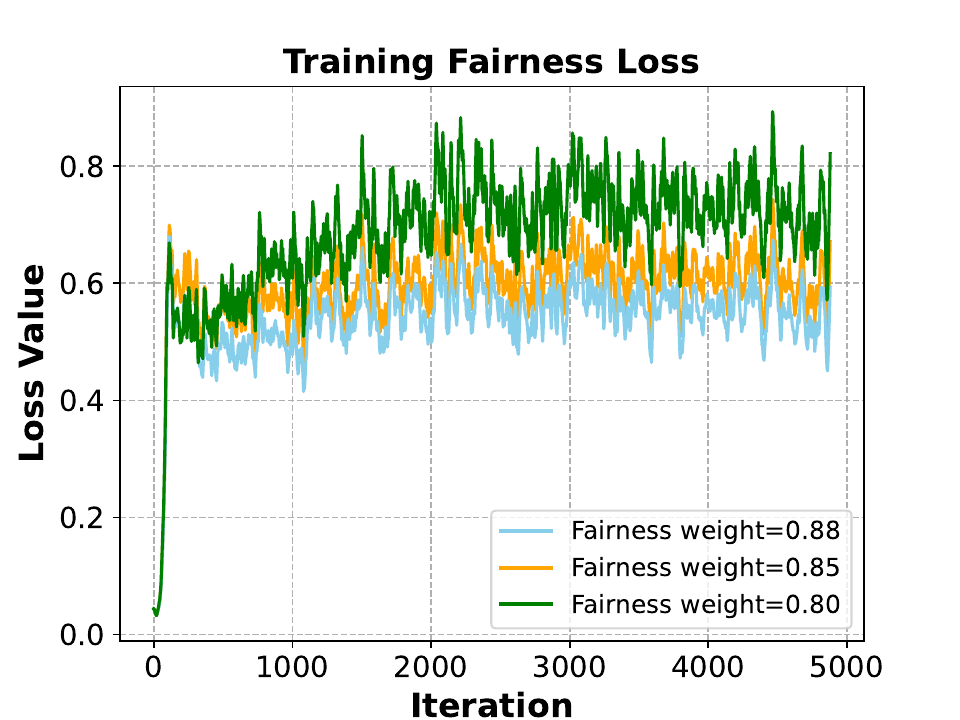}
    \caption{Fairness Loss}
  \end{subfigure}
  \caption{Moving average weights $\beta_{fair}\in\{0.80,0.85,0.88\}$ applied to the fairness gradients when using reference vector $v=(1,1)$. While the solution associated with $\beta_{fair}=0.85$ is the closest to $v$, increasing (decreasing) $\beta_{fair}$ introduces a bias further minimizing (maximizing) the fairness loss.\looseness-1}
  \label{fig:abstudy}
\end{figure}

We then explore how different moving average weights affect the optimization. 
We set reference vector to $\vec{v}=(1,1)$, fix the weight for accuracy ($\beta_{acc}=0.80$), and apply different weights ($\beta_{fair}=\{0.88, 0.85, 0.80\}$) for fairness. 
The results in Figure~\ref{fig:abstudy} indicate that increasing the moving average weight of one objective could make it more dominant in the optimization process. 
For example, when increasing $\beta_{fair}$ from 0.80 to 0.88, the fairness loss decreases and the classification loss increases accordingly. 
Although the model might be sensitive to the weights, it makes the training process more controllable.

\section{Conclusions}
\label{conclusion}

In this paper, we present CPT, a method for controllable Pareto fairness-accuracy trade-off. 
CPT provides two techniques to refine the application of the gradient-based multi-objective optimization method in fairness-accuracy trade-off. 
First, CPT applies moving average gradients instead of stochastic gradients for each objective, which stabilizes the training process and results in better fairness performance. 
Second, CPT generates a mask based on parameter magnitude to prune the gradient, the denoised low dimensional gradient benefits MOO by providing a more precise common descent vector. 
We evaluate CPT on real-world datasets and show its advantage in both optimization process and test results. 
In the future work, we would like to explore how to get a set of Pareto stationary solutions near the reference vector instead of a single solution for each vector. 

\section*{Limitations}

\paragraph{Sensitivity of moving average weights} 
Although applying moving average gradients can benefit the training process, it could be tedious to tune the moving average weights. 
And the weights may not always generalize well when training the model on various datasets due to the difference in data distribution. 

\paragraph{Trade-off within each class}
We have shown that CPT is able to generate controllable solutions based on the preference of fairness and accuracy over the whole training and test datasets. However, the performance in each class may not follow the preference as discussed in Appendix~\ref{sec: casestudy}.

\section*{Ethics Statement}

This work aims to provide diverse and controllable trade-offs between fairness and accuracy within the legal domain. We conduct experiments on the Jigsaw\footnote{https://www.kaggle.com/competitions/jigsaw-unintended-bias-in-toxicity-classification/data} and BiasBios\footnote{https://github.com/Microsoft/biosbias} datasets. Jigsaw dataset was publicized by Kaggle for the Toxic Comment Classification Challenge in 2017 and is already publicly available under CC BY-NC-SA 4.0 licenses. BiasBios dataset is also publicly available and has been widely used in fairness research work~\citep{ruhling2021end,han2022towards,shao2023erasure}. The safety and privacy issues have been checked by the original creators and we consider the data used in our experiment to be safe. 
As for the potential risk, the results showed in Table~\ref{tab:casestudy} indicate the difference in how much accuracy the model would sacrifice to achieve better fairness for each occupation. The reason could be data sampling or the task difficulty of each class is different.



\bibliography{acl/acl_latex}
\clearpage
\appendix

\section{Appendix}
\subsection{Computing Common Descent Vector}
\label{sec: common descent vector}
Considering the case of two objectives, the optimization problem could be defined as 
\begin{equation}
    \min _{\alpha \in[0,1]}\left\|\alpha \nabla_{\boldsymbol{{\bm{\theta}}}} {\mathcal{L}}_1\left(\boldsymbol{{\bm{\theta}}}\right)+
(1-\alpha) \nabla_{\boldsymbol{{\bm{\theta}}}} \hat{\mathcal{L}}_2\left(\boldsymbol{{\bm{\theta}}}\right)\right\|_2^2
\end{equation}
Then, the analytical solution for $\alpha$ is:
\begin{equation}
    \alpha=\frac{\left(\nabla_{\bm{\theta}} \mathcal{L}_2({\bm{\theta}})-\nabla_{\bm{\theta}} \mathcal{L}_1({\bm{\theta}})\right)^T * \nabla_{\bm{\theta}} \mathcal{L}_2({\bm{\theta}})}{\left\|\nabla_{\bm{\theta}} \mathcal{L}_1({\bm{\theta}})-\nabla_{\bm{\theta}} \mathcal{L}_2({\bm{\theta}})\right\|^2}
\end{equation}
When it comes to multiple objectives, the calculation of the common descent vector still relies on the inner product. 

\subsection{Hypervolume}
\label{sec: hypervolume}
Hypervolume is a valuable metric in multi-objective optimization that measures the quality of a set of solutions by quantifying the objective space they cover. 
The hypervolume metric can be defined as follows: 
given a set of points $P \subset \mathbb{R}^n$ and a reference point $\boldsymbol{r} \in \mathbb{R}_{+}^n$, the hypervolume of $\mathbb{R}$ is measured by the region of non-dominated points bounded above by $\boldsymbol{r}$: 
\begin{equation}
    H V(P)=\operatorname{VOL}\left(\left\{s \in \mathbb{R}_{+}^n \mid \exists p \in P:(p \preceq s) \wedge(s \preceq \boldsymbol{r}\right\}\right)
\end{equation}

\begin{figure*}
  \centering
  \begin{subfigure}{0.3\textwidth}
    \centering
    \includegraphics[width=\textwidth]{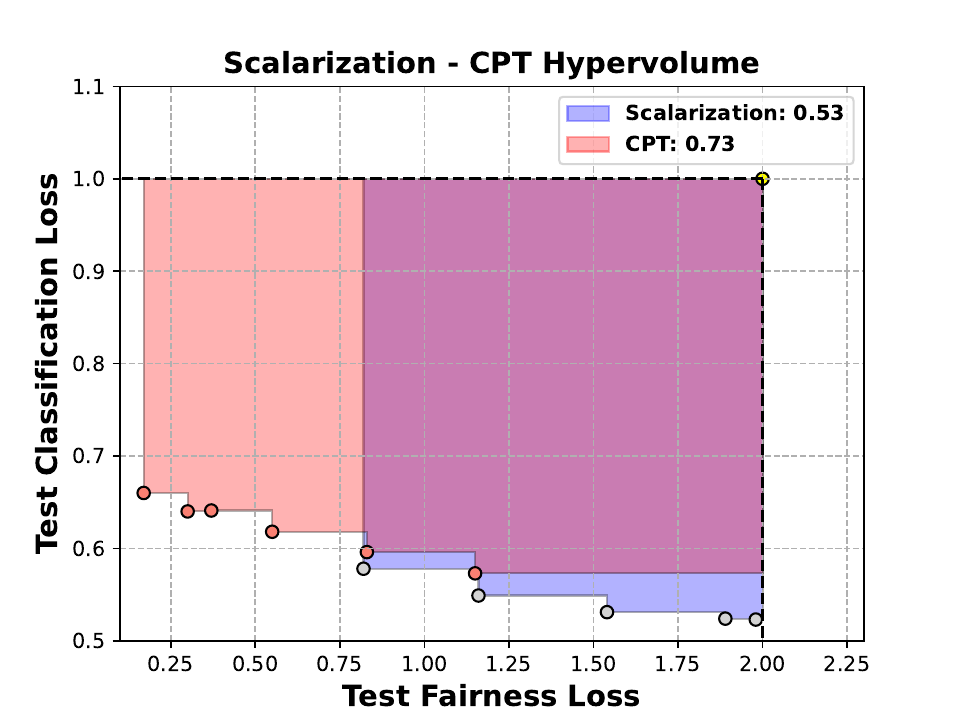}
    \caption{Scalarization-CPT(Jigsaw)}
  \end{subfigure}
  \begin{subfigure}{0.3\textwidth}
    \centering
    \includegraphics[width=\textwidth]{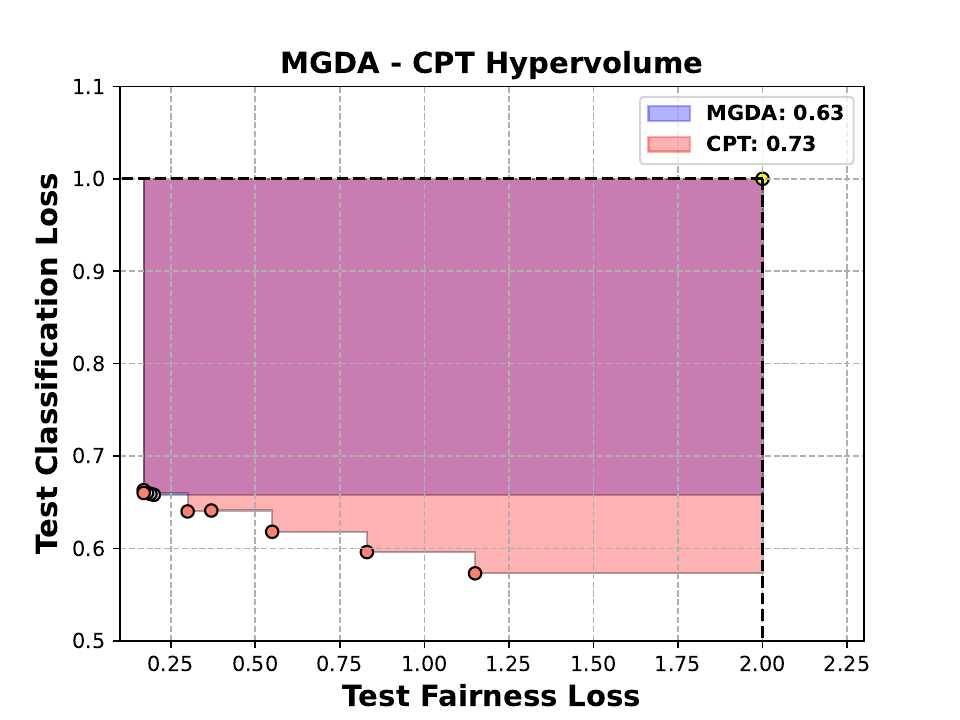}
    \caption{MGDA-CPT(Jigsaw)}
  \end{subfigure}
  \begin{subfigure}{0.3\textwidth}
    \centering
    \includegraphics[width=\textwidth]{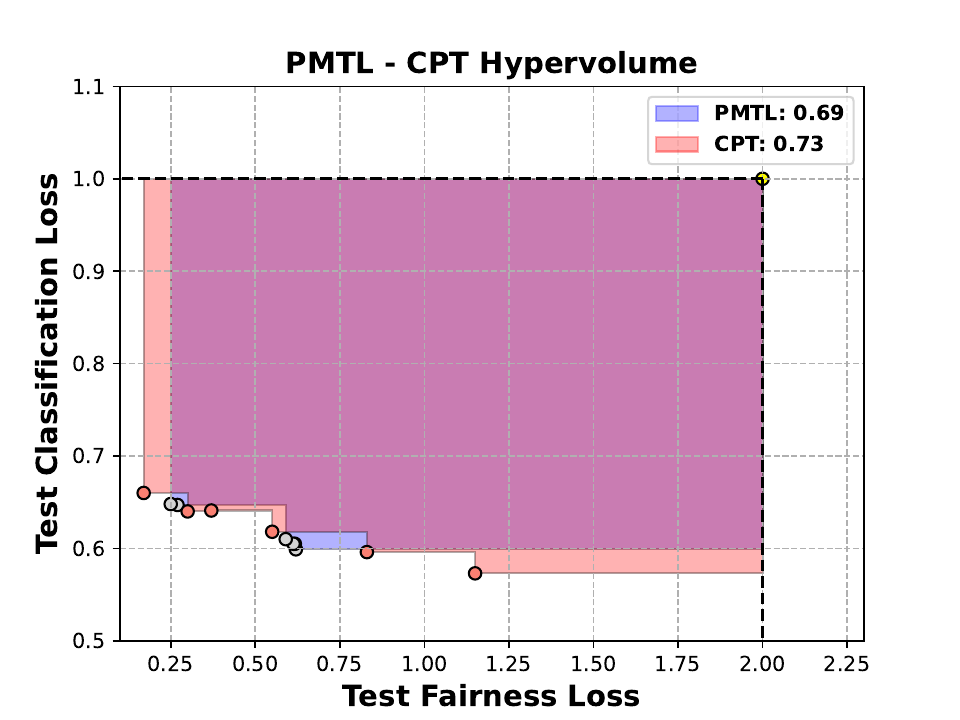}
    \caption{PMTL-CPT(Jigsaw)}
  \end{subfigure}
  \begin{subfigure}{0.3\textwidth}
    \centering
    \includegraphics[width=\textwidth]{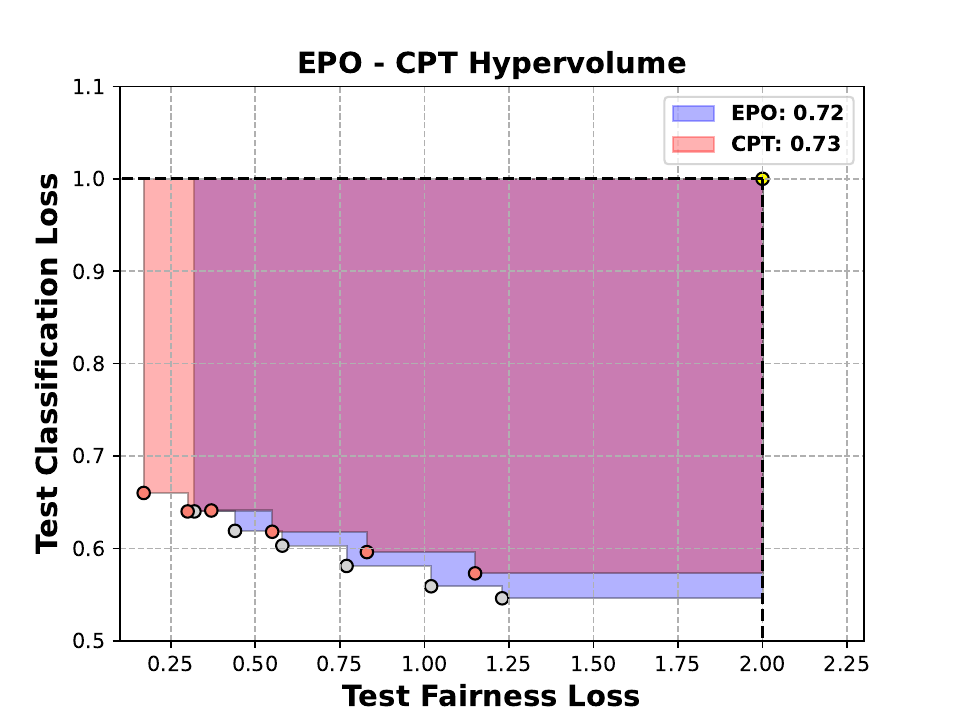}
    \caption{EPO-CPT(Jigsaw)}
    \label{fig:epo-cpt-j}
  \end{subfigure}
  \begin{subfigure}{0.3\textwidth}
    \centering
    \includegraphics[width=\textwidth]{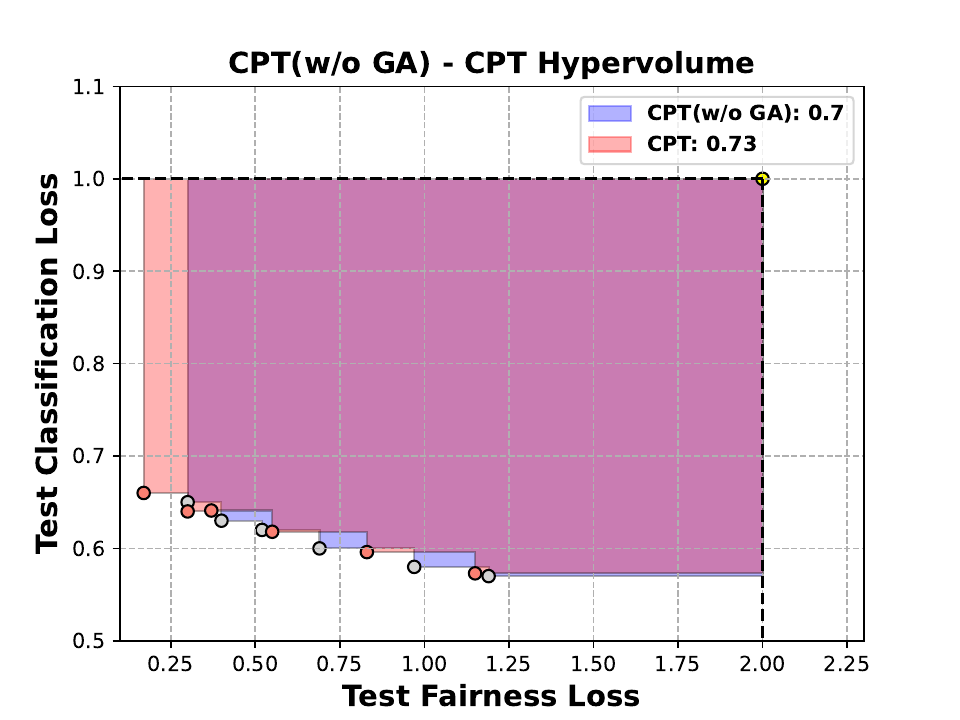}
    \caption{CPT(w/o GA)-CPT(Jigsaw)}
  \end{subfigure}
  \begin{subfigure}{0.3\textwidth}
    \centering
    \includegraphics[width=\textwidth]{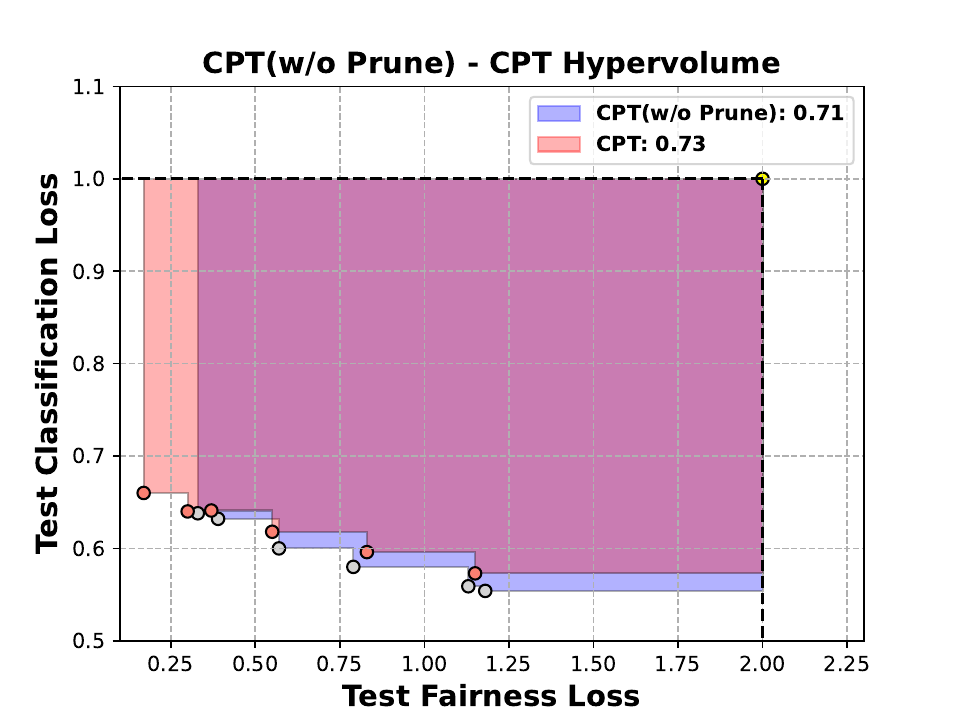}
    \caption{CPT(w/o Prune)-CPT(Jigsaw)}
  \end{subfigure}

   \begin{subfigure}{0.3\textwidth}
    \centering
    \includegraphics[width=\textwidth]{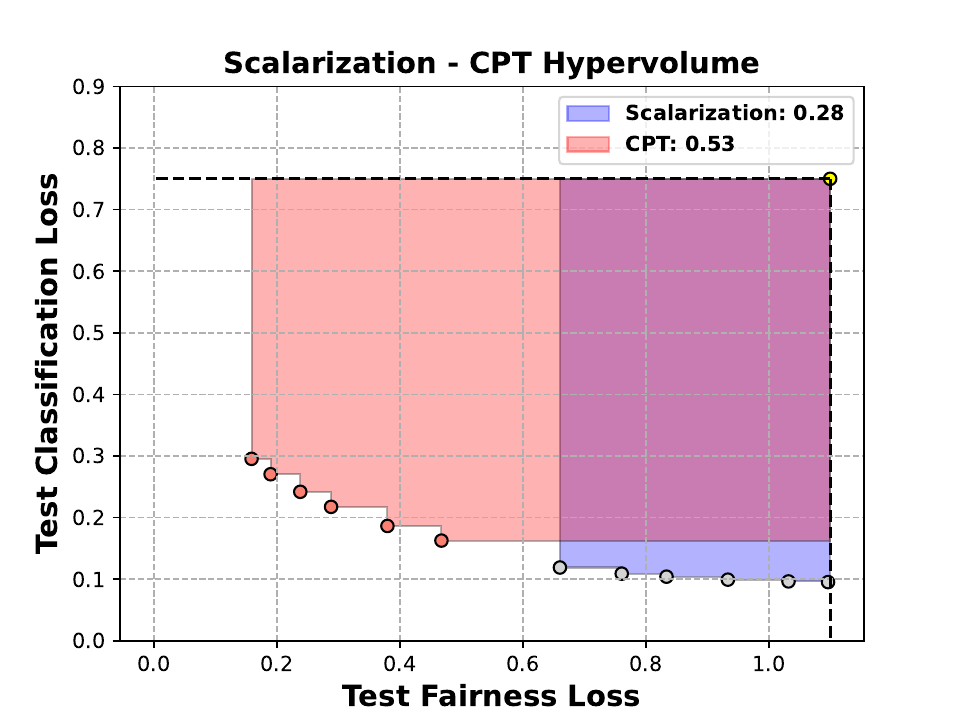}
    \caption{Scalarization-CPT(BiasBios)}
  \end{subfigure}
  \begin{subfigure}{0.3\textwidth}
    \centering
    \includegraphics[width=\textwidth]{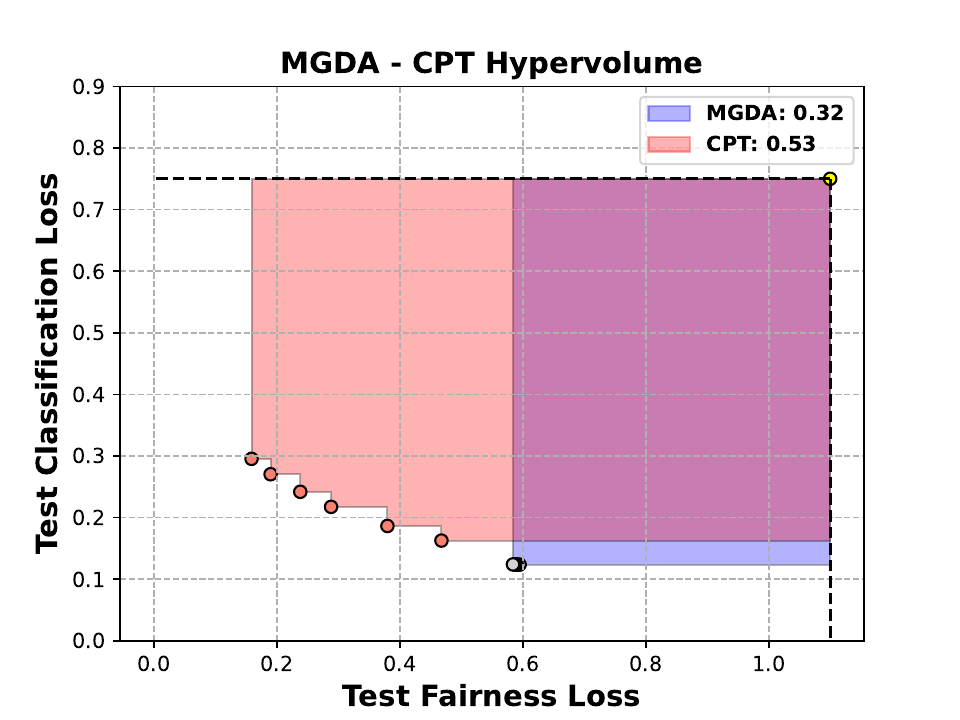}
    \caption{MGDA-CPT(BiasBios)}
  \end{subfigure}
  \begin{subfigure}{0.3\textwidth}
    \centering
    \includegraphics[width=\textwidth]{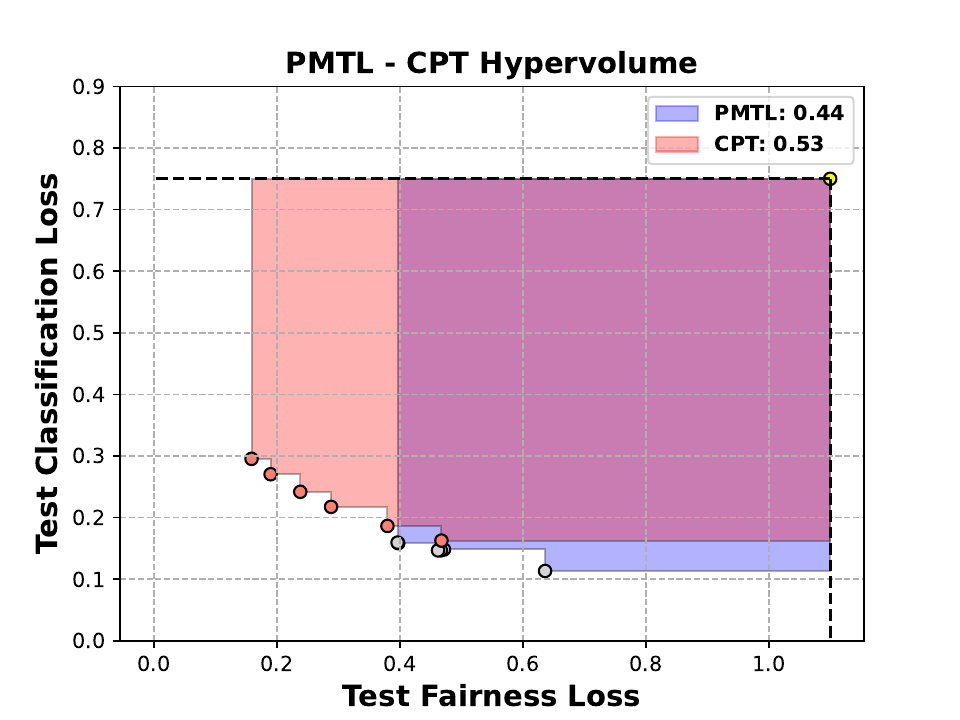}
    \caption{PMTL-CPT(BiasBios)}
  \end{subfigure}
  \begin{subfigure}{0.3\textwidth}
    \centering
    \includegraphics[width=\textwidth]{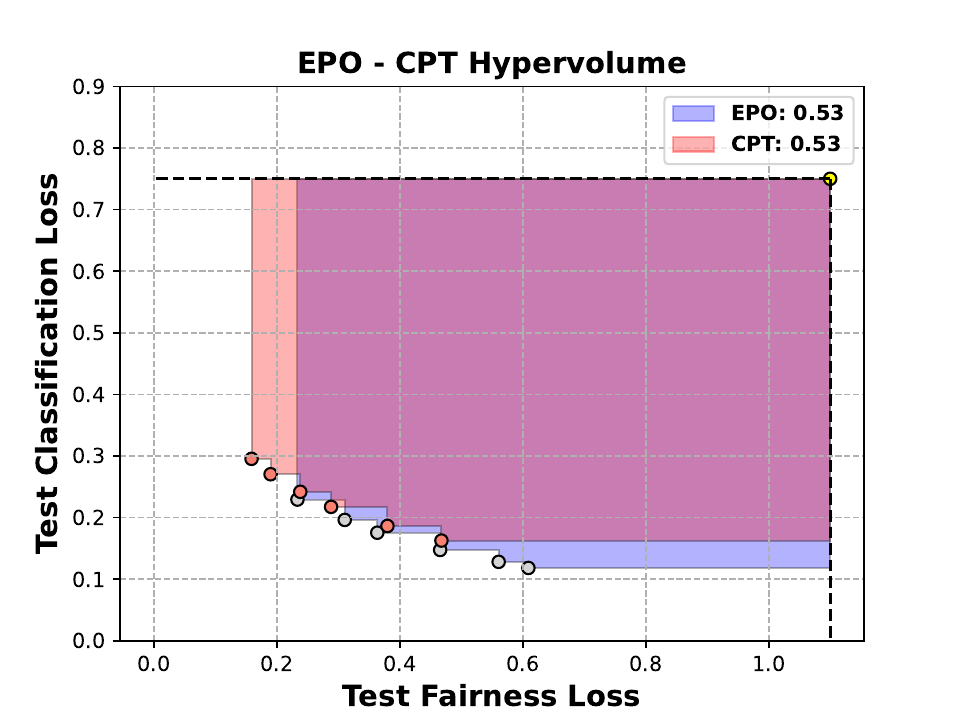}
    \caption{EPO-CPT(BiasBios)}
    \label{fig:hp(epo-cpt-bb)}
  \end{subfigure}
  \begin{subfigure}{0.3\textwidth}
    \centering
    \includegraphics[width=\textwidth]{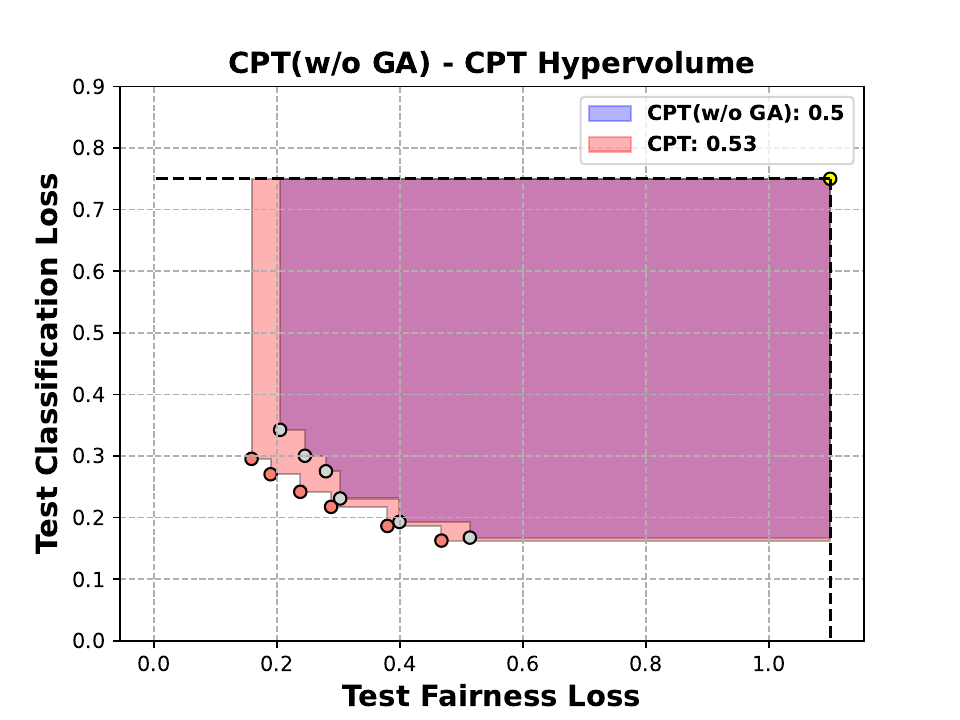}
    \caption{CPT(w/o GA)-CPT(BiasBios)}
  \end{subfigure}
  \begin{subfigure}{0.3\textwidth}
    \centering
    \includegraphics[width=\textwidth]{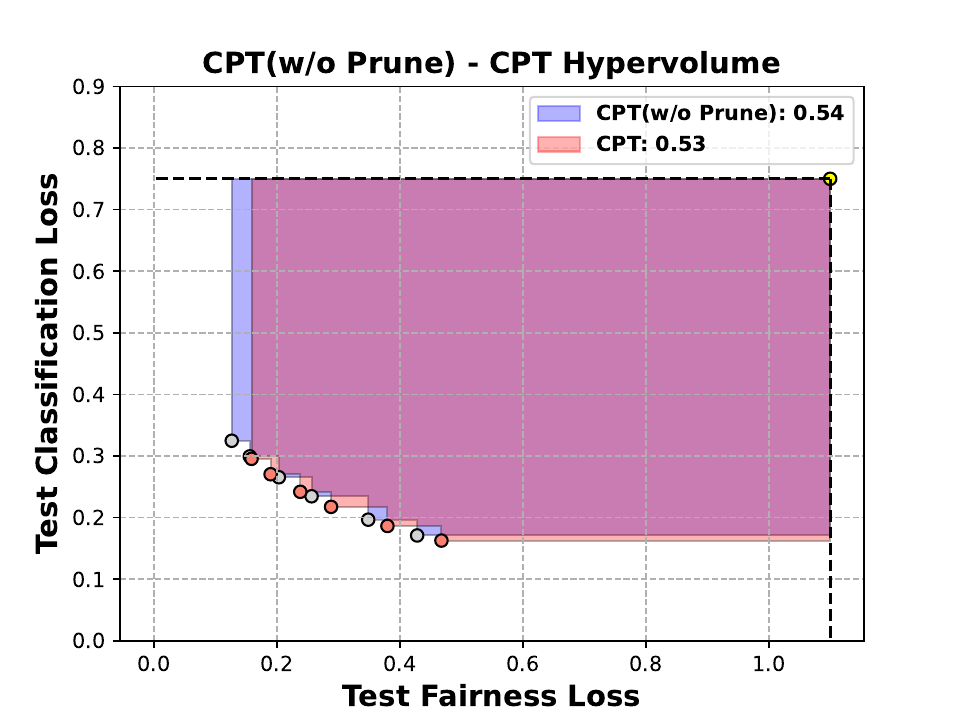}
    \caption{CPT(w/o Prune)-CPT(BiasBios)}
  \end{subfigure}

  \caption{Hypervolume (test set) of the solutions achieved by different methods in the fairness-accuracy space. Numerical results are reported in Table~\ref{tab:hp}. {\bf CPT achieves higher hypervolume, indicating the diversity of solutions} that provide different trade-offs.}
  \label{fig:hypervolume}
\end{figure*}

In the bi-optimization problem, it can be represented by the area of the polygon bounded by the solution set and reference point. We show the hypervolume on the test set in Figure~\ref{fig:hypervolume}.

\subsection{Case Study}
\label{sec: casestudy}
In order to showcase how different reference vectors can affect the model's performance, we conduct a case study on the BiasBios dataset. We first randomly sample 20 cases (10 for male and 10 for female) for each class.  Then we feed the samples into models trained with reference vector (2,1) and (1,2) and analyze the model's outputs. We repeat the whole process for three times and present the true positive rate (TPR) and accuracy in Table~\ref{tab:casestudy}. 

Model trained with (2,1) achieves better accuracy in most classes but there is a larger TPR gap between the two groups. 
Model trained with (1,2) tends to achieve better fairness by minimizing the TPR gap. But it does not apply to all classes, for "Psychologist" and "Dentist" classes, solutions may already satisfy Pareto stationary and there is no update on TPR and accuracy. For "Nurse" and "Surgeon" classes, model sacrifices accuracy to achieve better fairness. For "Physician" class, the TPR for the male group is improved, which leads to a smaller TPR gap as well as higher accuracy. 

It can be concluded that even though we achieve the controllable trade-offs on the whole dataset via different reference vectors, the preference may not always generalize to each class. The community may need to consider a fine-grained fairness loss function. Overall, the Pareto front of fairness and accuracy is still complicated and worth studying.

\begin{table}[]
    \small
    \centering
        \begin{tabular}{clcc}
            \toprule
            \multirow{2}{*}{\textbf{Class}}&
            \multirow{2}{*}{\textbf{Metric}}&
            \multicolumn{2}{c}{\textbf{Reference Vector}} \\
            \cmidrule{3-4}
            &&(2,1)&(1,2) \\
            \midrule
            & TPR(male) & 0.90 & 0.90 \\
            Psychologist& TPR(female) & 0.90 & 0.90 \\
            & Accuracy & 90.00 & 90.00\\
            \midrule
            & TPR(male) & 0.50 & 0.50 \\
            Nurse& TPR(female) & 0.80 & 0.70 \\
            & Accuracy & 65.00 & 60.00\\
            \midrule
            & TPR(male) & 0.50 & 0.10 \\
            Surgeon& TPR(female) & 0.40 & 0.10 \\
            & Accuracy & 45.00 & 10.00\\
            \midrule
            & TPR(male) & 1.00 & 1.00 \\
            Dentist& TPR(female) & 1.00 & 1.00 \\
            & Accuracy & 100.00 & 100.00\\
            \midrule
            & TPR(male) & 0.80 & 0.90 \\
            Physician& TPR(female) & 1.00 & 1.00 \\
            & Accuracy & 90.00 & 95.00\\
            \bottomrule
        \end{tabular}
    \caption{True positive rate (TPR) of two groups and accuracy with difference reference vectors. Comparing with model trained with (2,1), the gap between TPR of two groups is reduced when applying model trained with (1,2).\looseness-1}
    \label{tab:casestudy}
\end{table}

\subsection{Implementation Details}
\label{sec:train-details}
The version of Sentence Transformer we use is paraphrase-MiniLM-L3-v2 \footnote{https://huggingface.co/sentence-transformers/paraphrase-MiniLM-L3-v2}. 
The classifier consists of two fully connected layers with size (384, 384) and (384,1). 
We utilize SGD with 0.9 momentum. The learning rate is set to 0.01 initially and decreases every epoch with a 0.8 decay rate. 
The number of epochs is 40 and the batch size is set to 128. 
As for hyperparameters related to our method, we set the threshold $\psi$ to be 0.002. 

\subsection{Dataset Statistics}
\label{sec:data-stat}

Table~\ref{tab:jigsaw} shows the statistics of Jigsaw training set. For positive and negative classes, the data points for each race group are unbalanced. 
As for BiasBios training set (shown in in Table~\ref{tab:biasbios}), the distribution of the female group and male group in each occupation is also unbalanced. The imbalance in the training data brings more unfairness in the model's decision but sometimes could benefit the prediction accuracy, making them suitable datasets to study the trade-off between fairness and accuracy. 

\begin{table}[t]
    \small
    \centering
        \begin{tabular}{lll}
            \toprule
            \multirow{2}{*}{\textbf{Subgroup}}
             & \multicolumn{2}{c}{\textbf{Label}} \\
            \cmidrule{2-3}
            & Positive & Negative \\
            \midrule
            White & 5636 & 5410 \\
            Black & 3747 & 3050 \\
            Latino & 313 & 497\\
            Asian & 183 & 224 \\
            \bottomrule
        \end{tabular}
    \caption{Statistics of Jigsaw training-set.\looseness-1}
    \label{tab:jigsaw}
\end{table}

\begin{table}[ht]
    \small
    \centering
    \begin{tabular}{lll}
        \toprule
        \multirow{2}{*}{\textbf{Occupation}}
         & \multicolumn{2}{c}{\textbf{Gender}} \\
        \cmidrule{2-3}
        & Female & Male \\
        \midrule
        Psychologist & 7491 & 4400 \\
        Surgeon & 1203 & 7424 \\
        Nurse & 11178 & 1153\\
        Dentist & 3283 & 6128 \\
        Physician & 10782 & 14285 \\
        \bottomrule
    \end{tabular}
    \caption{Statistics of BiasBios training-set.\looseness-1}
    \label{tab:biasbios}
\end{table}
\begin{figure}[ht]
    \centering
    \includegraphics[width=0.45\textwidth]
    {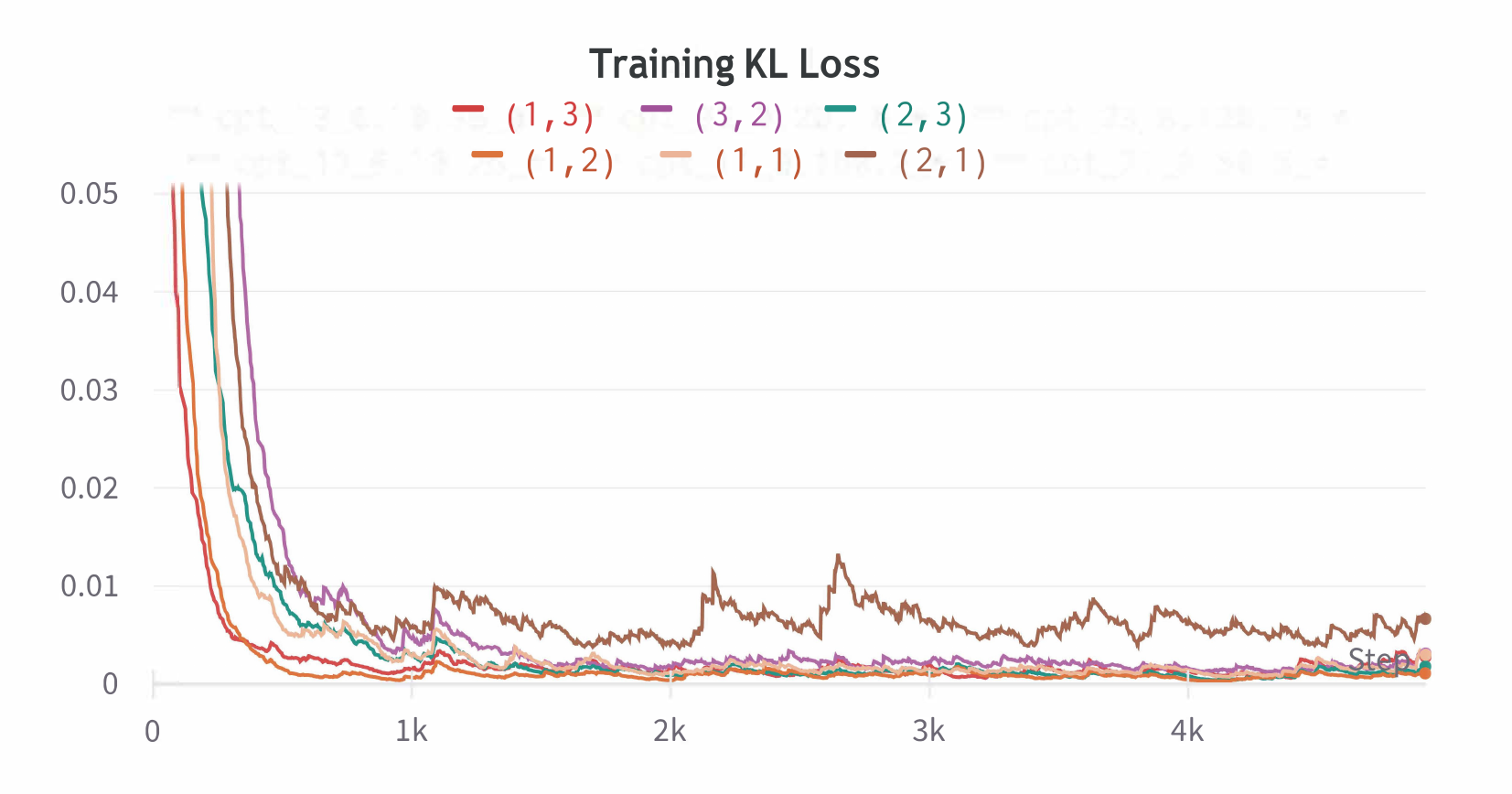}
    \caption{Training KL divergence loss: The KL loss decreases from correction stage to MOO stage and converges at the end of the training, which indicates the optimization process follows the reference vector very well.}
    \label{fig:training-kl-loss}
\end{figure}

\end{document}

%% file: math_commands.tex
\usepackage{amsmath,amsfonts,bm}

\def\eqref#1{equation~\ref{#1}}

\def\1{\bm{1}}

\DeclareMathAlphabet{\mathsfit}{\encodingdefault}{\sfdefault}{m}{sl}
\SetMathAlphabet{\mathsfit}{bold}{\encodingdefault}{\sfdefault}{bx}{n}

\def\sV{{\mathbb{V}}}

